%% file: main.tex
\documentclass[10pt,twocolumn,letterpaper]{article}

\usepackage{cvpr}              


\definecolor{cvprblue}{rgb}{0.21,0.49,0.74}
\usepackage[pagebackref,breaklinks,colorlinks,allcolors=cvprblue]{hyperref}

\usepackage{graphicx}
\usepackage{amsmath}
\usepackage{amssymb}
\usepackage{booktabs}
\usepackage{multirow}
\usepackage{algorithm}
\usepackage{algorithmicx}
\usepackage{algpseudocode}
\usepackage{color}

\def\paperID{10733} 
\def\confName{CVPR}
\def\confYear{2025}

\title{Improving Adversarial Transferability on Vision Transformers\\via Forward Propagation Refinement}

\author{Yuchen Ren$^{1}$, Zhengyu Zhao$^{1}$\thanks{Corresponding author}, Chenhao Lin$^{1}$, Bo Yang$^{2}$, Lu Zhou$^{3}$, Zhe Liu$^{4}$, Chao Shen$^{1}$\\
$^{1}$Xi'an Jiaotong University, China; $^{2}$Information Engineering University, China;\\
$^{3}$Nanjing University of Aeronautics and Astronautics, China; $^{4}$Zhejiang Lab, China\\
{\tt\small ryc98@stu.xjtu.edu.cn, }{\tt\small \{zhengyu.zhao, linchenhao\}@xjtu.edu.cn, }\\
{\tt\small yangbo\_hn@163.com, }{\tt\small \{lu.zhou, zhe.liu\}@nuaa.edu.cn, chaoshen@mail.xjtu.edu.cn}
}

\begin{document}
\maketitle

\begin{abstract}

Vision Transformers (ViTs) have been widely applied in various computer vision and vision-language tasks. 
To gain insights into their robustness in practical scenarios, transferable adversarial examples on ViTs have been extensively studied. 
A typical approach to improving adversarial transferability is by refining the surrogate model. 
However, existing work on ViTs has restricted their surrogate refinement to backward propagation. 
In this work, we instead focus on Forward Propagation Refinement (FPR) and specifically refine two key modules of ViTs: attention maps and token embeddings. 
For attention maps, we propose Attention Map Diversification (AMD), which diversifies certain attention maps and also implicitly imposes beneficial gradient vanishing during backward propagation. 
For token embeddings, we propose Momentum Token Embedding (MTE), which accumulates historical token embeddings to stabilize the forward updates in both the Attention and MLP blocks. 
We conduct extensive experiments with adversarial examples transferred from ViTs to various CNNs and ViTs, demonstrating that our FPR outperforms the current best (backward) surrogate refinement by up to 7.0\% on average.
We also validate its superiority against popular defenses and its compatibility with other transfer methods.
Codes and appendix are available at \url{https://github.com/RYC-98/FPR}.

\end{abstract}

\section{Introduction}
\label{sec1}

Deep neural networks (DNNs) \cite{resnet,resnext,wang1} demonstrate remarkable performance across diverse domains, with Vision Transformers (ViTs) \cite{vit,deit,pit} excelling particularly in computer visualization.
However, ViTs are also shown to be vulnerable to adversarial examples \cite{v_fragile,v_intriguing}, which are optimized by adding subtle perturbations \cite{epca,perc} into clean images~\cite{fgsm,logit,yang}.

A key property of adversarial examples that causes practical security concerns is their transferability, i.e., they can be optimized on a (seen) surrogate model and then directly transfer their effects to (unseen) target models~\cite{di,mi}. 
Compared to other transfer approaches that focus on loss design, optimization algorithm or data augmentation~\cite{vt,sfva,mumodig}, surrogate refinement~\cite{sgm,ghost,bpa} explores the intrinsic relationship between the model itself and adversarial transferability, offering a unique perspective.

\begin{figure}[t]
\centering
    \includegraphics[width=0.95\columnwidth]{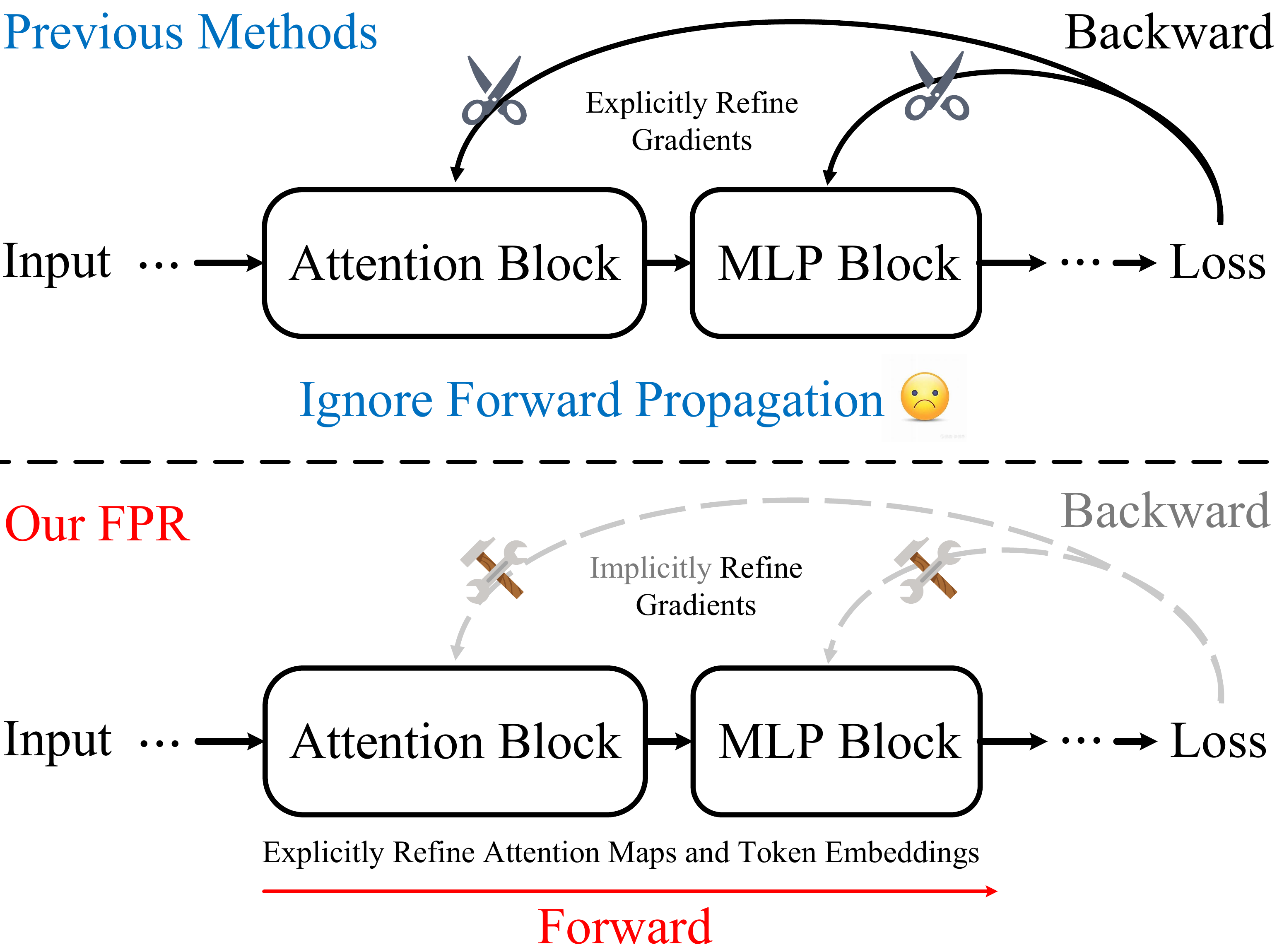}
\caption{Our Forward Propagation Refinement (FPR) vs. previous methods which only refine the backward propagation.}
\label{dif}
\end{figure}

However, existing surrogate refinements on ViTs~\cite{pna,tgr} are limited to refining the backward propagation, i.e., only modifying the gradient information.
To address the above limitation, in this work, we explore the potential of Forward Propagation Refinement (FPR), i.e., modifying features during forward propagation.
Our FPR not only provides a new perspective of surrogate refinement but also potentially benefits the refinement during the subsequent backward propagation (see our discussion in Section~\ref{amdsec}). 
Specifically, FPR comprises two parts that refine two key components of ViTs respectively: attention maps and output token embeddings. Figure \ref{dif} highlights the difference between our forward-based method and the previous backward-based methods.

For attention maps, we identify that certain attention maps in ViTs are redundant and can deteriorate transferability.
Rather than simply dropping redundant attention maps, which recklessly removes all information, we propose Attention Map Diversification (AMD), which moderately diversifies the attention maps based on random weighting. The introduced randomness on attention maps acts as a simulation of unknown models, thereby mitigating the overfitting to the surrogate model. Although AMD only modifies the forward propagation, we find that it can potentially induce the vanishing of certain gradients flowing through the attention maps during backward propagation, further alleviating overfitting \cite{pna}.

For output token embeddings, we posit that adversarial examples during generation are highly likely to be near the local optima, resulting in the surrogate model's embeddings containing erroneous information. Consequently, inspired by the momentum distillation technology when training the vision-language pretraining model \cite{albef}, we propose Momentum Token Embedding (MTE), to stabilize the updates of token embeddings in each iteration by accumulating historical token embedding information. 

In sum, we make the following main contributions:
\begin{itemize}

\item We explore the refinement of ViT surrogate models from the perspective of forward propagation, particularly modifying the attention maps and output token embeddings.

    \item We propose Attention Map Diversification (AMD) to moderately avoid attention redundancy, which also induces beneficial gradient vanishing during the subsequent backward propagation, altogether mitigating the overfitting to the surrogate model.

    \item We propose Momentum Token Embedding (MTE) to stabilize the updates of token embeddings by accumulating historical information, during the generation of adversarial examples on ViTs.

    \item Extensive experiments validate that our Forward Propagation Refinement (FPR), comprising AMD and MTE, outperforms the current best (backward) refinement method by up to an average of 7.0\% against 13 ViTs and CNNs with or without defenses.
\end{itemize}

\section{Related Work}

\subsection{Surrogate Refinement Attacks}
In this subsection, we specifically delineate attack methods falling within the surrogate refinement, since it is the main topic of this work. Skip Gradient Method (SGM) \cite{sgm} attenuates gradients traversing the residual modules to mitigate overfitting to the surrogate model. Ghost Networks \cite{ghost} employs dropout in each block of the surrogate model and perturbs the skip connections to create a longitudinal ensemble. Backward Propagation Attack (BPA) \cite{bpa} demonstrates that gradients traversing ReLU and max-pooling layers can carry inaccurate information. To address this, it adopts both non-monotonic function and softmax with temperature to recover the true gradients, thereby enhancing transferability. 
The aforementioned methods focus on the backward propagation of CNNs, while a few approaches also modify the output features of convolutional layers during forward propagation.
Diversifying the High-level Features (DHF) \cite{dhf} modifies the high-level features through random transformations.
Clean Feature Mixup (CFM) \cite{cfm} introduces competition into the generation of adversarial examples at the feature level for stronger transferability. 
Several methods also retrain the surrogate model to strengthen the transferability \cite{dra,bay}, however, this may significantly increase the attack cost. 

The above surrogate refinements are primarily designed for CNNs; several methods have also been developed for ViTs. Pay No Attention and Patch Out (PNAPO) \cite{pna} comprises three modifications to synergistically mitigate overfitting to the surrogate model. First, it cuts off gradients traversing all attention maps. Second, it randomly updates a specific number of input patches. Finally, it adds an extra term to the cross-entropy loss function that increases the adversarial noise norm to enhance transferability. Token Gradient Regularization (TGR) \cite{tgr} posits that extreme values in gradients are detrimental. Therefore, it removes the extreme gradients of QKV embeddings, attention maps, and MLP blocks to reduce gradient variance, thereby alleviating overfitting to the surrogate models. Gradient Normalization Scaling (GNS) \cite{gns} improves TGR by accurately normalizing and scaling the identified mild gradients, which yields stronger transferability.
Even though the above methods exhibit the ability to craft transferable adversarial examples, they neglect the broader implications of forward propagation, which prevents further transferability. In this work, we focus on refining the ViT surrogate model from the new perspective of forward propagation and specifically explore two key components: attention maps and output token embeddings.

\begin{figure*}[!t]
\centering
\includegraphics[width=1.9\columnwidth]{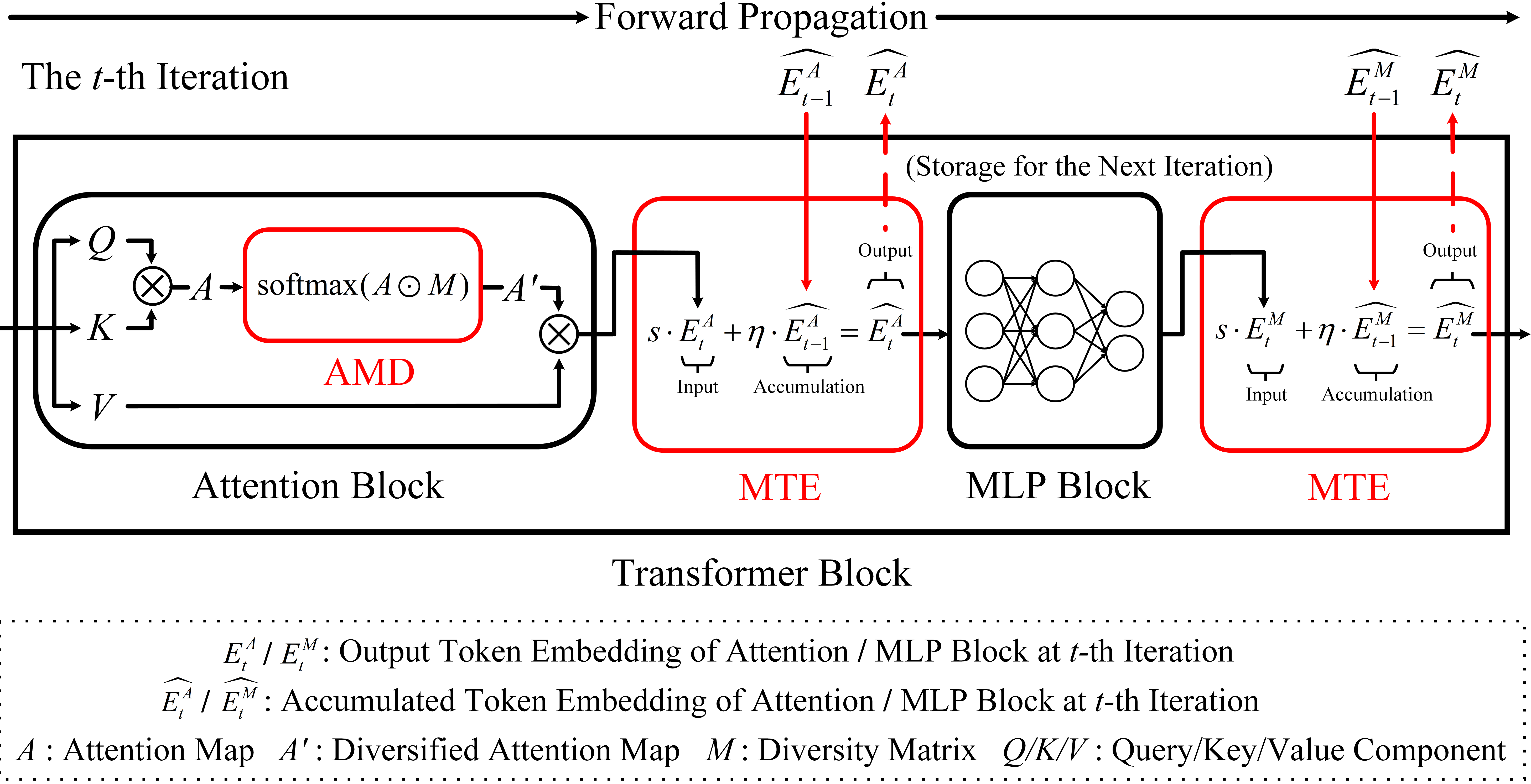}
\caption{The overview of Forward Propagation Refinement (FPR). Note that Attention Map Diversification (AMD) is applied to specific attention maps according to the index set, while Momentum Token Embedding (MTE) is applied to all Attention and MLP blocks.}
\label{overview}
\end{figure*}

\subsection{Other Transfer-based Attacks}
This subsection reviews transfer-based attacks \cite{revist} that do not need to modify the surrogate models. Fast Gradient Sign Method (FGSM) \cite{fgsm} creates adversarial examples in a single step by following the gradient sign's direction. Basic Iterative Method (BIM) \cite{bim} extends FGSM by applying multiple iterations. Momentum Iterative Method (MIM) \cite{mi} introduces a momentum term for gradients at each iteration for better convergence. Nesterov Fast Gradient Sign Method (NIM) \cite{ni} takes an extra step at the beginning of each iteration and adopts the example at the farther location to compute the gradients for the update. Variance Tuning Method (VTM) \cite{vt} modifies current gradients with variance at the last iteration. Gradient Relevance Attack (GRA) \cite{gra} utilizes correlation information extracted from the nearby samples to stabilize the current update direction. Some input transformations have also been presented to enhance transferability. Diversity Input Method (DIM) \cite{di} combines padding and resizing transformations on the input to reduce overfitting. Scale Invariance Method (SIM) \cite{ni} employs a group of scaled copies to enhance the input. Structure Invariant Attack (SIA) \cite{sia} employs different transformations from a library to each block of the input to prompt the transferability. Block Shuffle and Rotation (BSR) \cite{bsr} divides the input image into several blocks, which are then randomly shuffled and rotated to create a set of new images for gradient computation.

\subsection{Vision Transformers}
Transformer \cite{transformer} is widely used in natural language processing and has since gained increasing recognition in computer vision. ViT \cite{vit} processes images by dividing them into patches and embedding these patches as tokens. They employ a Transformer architecture with multi-head self-attention mechanisms to effectively capture spatial relationships. DeiT \cite{deit} improves ViT through teacher-student distillation, enabling superior performance with less data. Swin \cite{swin} employs a hierarchical structure with shifted windows to efficiently handle multi-scale attention. PiT \cite{pit} utilizes pooling layers to streamline computational processes and manage data flow. CaiT \cite{cait} incorporates LayerScale to stabilize training as model depth increases and employs Class-Attention layers to enhance the extraction of classification information. Visformer \cite{visf} combines convolutional layers with ViT to effectively integrate spatial features and attention mechanisms. CoaT \cite{coat} merges convolution and self-attention in a hierarchical architecture to balance feature extraction and feature interaction.

\section{Methodology}
In this section, we propose the Forward Propagation Refinement (FPR) to enhance the transferability of adversarial examples from ViTs, consisting of Attention Map Diversification (AMD) and Momentum Token Embedding (MTE). Figure \ref{overview} provides an overview of FPR.


\subsection{Preliminary}

Assume an image $x$ with channel $C$, height $H$, width $W$, the true label $y$, and a surrogate model $F$, the goal of a transfer-based attack is to add adversarial perturbation $\delta$ to the input $x$, creating an adversarial example ${x^a} = x + \delta$, which is then fed into the target model ${F^t}$ to induce incorrect predictions. This process can be formulated as solving the following problem:

\begin{equation}
    \label{goal}
    \delta ' = \mathop {\arg \max }\limits_\delta  L(F(x + \delta ),y), \,\\
    s.t. \,\, {\left\| \delta  \right\|_\infty } \le \epsilon,
\end{equation}
where $L$ is the loss function and $\epsilon$ is the maximum perturbation bound.

Before the image $x$ is fed into ViTs, it is segmented into $N = \frac{H \cdot W}{{P^2}}$ patches, with each patch having a height, width, and channel dimension of $P$, $P$, and $C$, respectively. Then, these patches are converted into token embeddings, and fed into a sequence of Transformer blocks. Each Transformer block includes an Attention block and a MLP block.

The input token embedding of the Attention block is transformed into Query ($Q$), Key ($K$), and Value ($V$) components.
Here, the shapes of $Q$, $K$ and $V$ are $N \times D$, where $D$ is the embedding dimension. The Attention block employs the self-attention mechanism to establish the relationship between $Q$, $K$, and $V$. The attention map $A$ in the Attention block is computed by:
\begin{equation}
\label{plain_attention_map}
A = {\rm{softmax(}}\frac{{Q{K^T}}}{{\sqrt D }}),
\end{equation}
where the softmax function is employed at each row. The attention map $A$ is then used to weight the $V$, and the output $Z$ is calculated as $Z = AV$.

The MLP block follows the Attention block and typically includes the fully connected layer, GELU activation function, dropout layer, and identity normalization layer to aggregate information for all token embeddings.

\begin{figure}[!t]
\centering
\includegraphics[width=0.8\columnwidth]{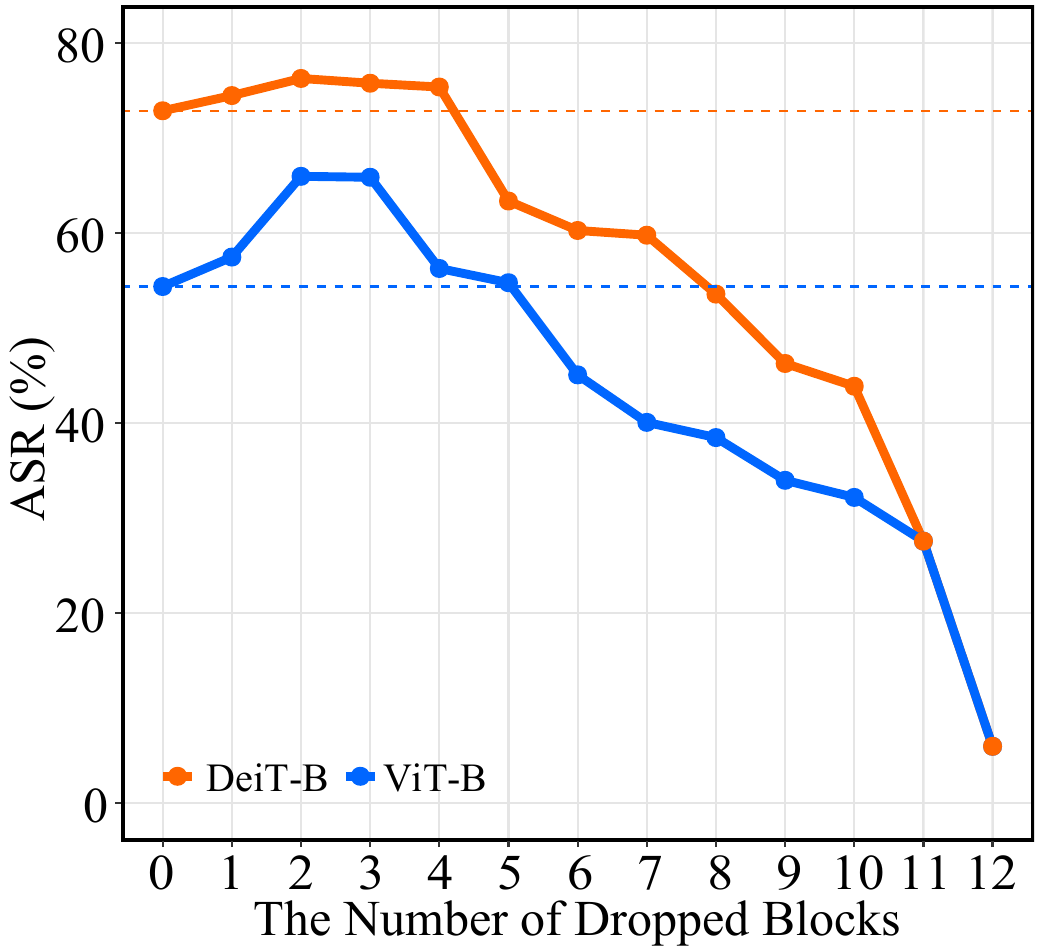}
\caption{Illustration of redundancy in attention maps: Randomly dropping a small number of blocks leads to improved transferability.
The results are averaged over 5 repeated runs on 13 target models, 
and the dash lines denote the baseline results of the MIM \cite{mi} without dropping.}
\label{motivation}
\end{figure}

\subsection{Attention Map Diversification (AMD)}
\label{amdsec}

The previous work \cite{v_redundancy} has demonstrated the redundancy in attention blocks of ViTs, showing that dropping a few such blocks does not significantly degrade classification accuracy. Therefore, it is reasonable to posit that such redundancy may also exist in generating adversarial examples, causing potential overfitting to the surrogate model.
To verify this, we conduct exploratory experiments by dropping certain attention blocks.
As can be seen from Figure \ref{motivation}, dropping a small number of blocks indeed leads to improved transferability (while dropping too many may harm transferability).

To avoid such redundancy, a straightforward idea is to apply dropout of attention blocks while generating adversarial examples.
However, we find that the dropout is too aggressive and may cause serious information loss, leading to only a small improvement of transferability (see comparing results in Table~\ref{ablations} of Section~\ref{ablsec}).
Therefore, we propose a more moderate method called Attention Map Diversification (AMD), which transforms the original attention map $A$ by multiplying it with a diversity matrix $M$.
Specifically, the diversified attention map $A'$ is obtained by:

\begin{equation}
\label{amd}
A' = AMD(A) = {\rm{softmax}}(A \odot M),
\end{equation}
where $\odot$ is the element-wise product and softmax function is applied to each row.
The diversity matrix $M$ has the same shape as $A$, and every element ${M_{i,j}}$ ($1 \le i,j \le N$) conforms to the uniform distribution $U(1 - d,1 + d)$, where $d$ is a factor that controls the diversity level. 
Note that AMD is applied to attention maps at specified Transformer blocks according to a index set $I$.

\noindent\textbf{The benefit of AMD on backward propagation.} In principle, modifying the forward propagation would impact the subsequent backward propagation. Here, we theoretically demonstrate that the impact of our AMD on backward propagation can further improve transferability.
Specifically, we find that AMD implicitly induces gradient vanishing at attention map positions during backward propagation, which aligns with a previous surrogate refinement, PNAPO ~\cite{pna}, that directly truncates the gradients of attention maps during backpropagation. The underlying principle of PNAPO suggests that gradients passing through attention maps carry overfitting information from surrogate models, and attenuating these gradients can enhance the transferability of adversarial examples.
Our findings can be explained as follows.

When incorporating AMD, we let $S = A \odot M$ and the $i$-th row vector of diversified attention map ${A'}_i$ as:
\begin{equation}
\begin{array}{ll}
{{A'}_i} & = ({{A'}_i}_{,1},{{A'}_i}_{,2},...,{{A'}_i}_{,N}) \\
         & = {\rm{softmax}}({A_{i,1}} \cdot {M_{i,1}},{A_{i,2}} \cdot {M_{i,2}},...,{A_{i,N}} \cdot {M_{i,N}}) \\
         & = {\rm{softmax}}({S_{i,1}},{S_{i,2}},...,{S_{i,N}})
\end{array}
\end{equation}
and the derivative $\frac{{\partial {{A'}_{i,m}}}}{{\partial {S_{i,n}}}}$ ($1 \le m,n \le N$) is expressed by:
\begin{equation}
\label{deritative}
\frac{{\partial {{A'}_{i,m}}}}{{\partial {S_{i,n}}}} = \left\{
\begin{array}{ll}
{{{A'}_{i,m}} \cdot (1 - {{A'}_{i,m}}),} & \text{if } m = n \\
{ - {{A'}_{i,m}} \cdot {{A'}_{i,n}},} & \text{if } m \ne n
\end{array}
\right.
\end{equation}

As the diversity factor $d$ becomes large, elements in the diversity matrix $M$ will fluctuate significantly, causing extreme values in $S$, and ultimately pushing the elements in $A'$ towards 0 or 1 due to the property of softmax function. From Eq. \ref{deritative}, whether the elements in $A'$ approach 0 or 1, the derivative will tend towards 0, resulting in gradient vanishing. According to the chain rule of differentiation, the gradients of the attention maps subjected to our AMD will experience varying degrees of gradient vanishing. 


Compared with PNAPO, which directly discards all gradients flowing through the attention maps, the gradient vanishing in our AMD is implicit without directly altering any aspect of the backward propagation. Furthermore, the gradient vanishing in our AMD method is more fine-grained, occurring at the row level of the attention map. Each row experiences varying degrees of vanishing, introducing additional randomness that helps to further mitigate overfitting to the surrogate model.

\subsection{Momentum Token Embedding (MTE)}
It is challenging to ensure convergence to the global optimum in the high-dimension space when generating adversarial examples. Typically, adversarial examples converge to a local optimum along an inaccurate path \cite{mi}. This causes the internal token embeddings of the surrogate model to contain inaccurate information during iterations, thereby weakening the transferability. 
Therefore, we borrow the idea of Momentum Distillation (MoD) \cite{albef} from the Vision-and-Language Pre-training (VLP) literature, where a momentum version of the model through the moving average of the primary model's parameters is maintained to mitigate the noise in the web-sourced training dataset.
Similarly, we maintain the momentum version of token embedding updates to accumulate historical token embeddings while generating adversarial examples.

We term this method Momentum Token Embedding (MTE), and formulate it as follows:
\begin{equation}
\label{mte}
\widehat {{E_t}} = MTE({E_t}) = \eta  \cdot \widehat {{E_{t - 1}}} + s \cdot {E_t},
\end{equation}
where $t$ is the iteration index, $E_t$ is the output token embedding of any Attention block or MLP block at the $t$-th iteration, $\widehat {{E_t}}$ is the corresponding accumulated token embedding at the $t$-th iteration, $\eta$ is the attenuation factor, and $s$ is the scale factor. MTE is applied to all Attention and MLP blocks.
Note that, different from MoD, our MTE does not require model training and can be seen as constructing a residual structure for output token embeddings across the time dimension.
Previous work also finds the usefulness of the residual structure in improving transferability, but during backward propagation within each iteration~\cite{sgm}. 

\begin{algorithm}[!htb]
\caption{Forward Propagation Refinement (FPR)}
{\bf Input:}
A surrogate model ${F}$ with the cross-entropy loss ${L }$, a clean image ${x}$  and its true label ${y}$. The maximum perturbation bound $\epsilon$, the iteration number $T$, the decay factor $\mu$, step size $\alpha$, index set $I$, diversity factor $d$, scale factor $s$ and attenuation factor $\eta$.\\
{\bf Output:}
The adversarial image $x^{adv}=x_{T}$.
\begin{algorithmic}[1]
 \State \textbf{Initialize} ${g_0} = 0$, $E_{0}=0$, $x_0 = x$.
    \For{$t = 1 \to T$}
      \Statex \textbf{During the forward propagation:}
        \State Apply AMD (Eq. (\ref{amd})) to attention maps in specific Transformer blocks according to the index set $I$.
        \State Apply MTE (Eq. (\ref{mte})) to output token embeddings in all Transformer (Attention and MLP) blocks.        

      \Statex \textbf{During the backward propagation:}
      \State Calculate the gradient: ${\nabla _{{x_{t-1}}}}L(F({x_{t-1}}),y)$
      \State Update momentum gradient: 
      $${g_t} = \mu  \cdot {g_{t - 1}} + \frac{{{\nabla _{{x_{t-1}}}}L(F({x_{t-1}}),y)}}{{{{\left\| {{\nabla _{{x_{t-1}}}}L(F({x_{t-1}}),y)} \right\|}_1}}}$$
      \State Update the adversarial example:
      $${x_t} = Cli{p_{\epsilon}}\{ x_{t - 1} + \alpha  \cdot sign({g_t})\} $$
    \EndFor
\end{algorithmic}
\label{al:1}
\end{algorithm}

\subsection{Complementary AMD and MTE}
The previous subsections examine AMD and MTE from their respective perspectives. 
Here, we further discuss why they can collaboratively enhance transferability. 

Specifically, AMD and MTE refine the forward propagation of ViTs from two complementary angles. 
AMD falls under the category of ``external refinement'', which enhances surrogate models by introducing extraneous information or perturbing existing embeddings. 
Techniques such as attention map dropout also fall into this category. Our empirical results demonstrate that this type of refinement is relatively aggressive, making it unsuitable for concurrent use or for application across all blocks.
In contrast, MTE belongs to the ``internal refinement'' category, mainly leveraging information from the input data itself.
Since these refinements rely on self-related information, they are typically more moderate and can be applied to all blocks.
Both AMD and MTE help promoting transferability but in different ways, this complementary effect further enhances the transferability of adversarial examples.

Finally, we provide the pseudocode of FPR (based on MIM \cite{mi}) in Algorithm \ref{al:1}. Specifically, at each of $T$ iterations, we first apply AMD to specific attention maps according to the index set $I$ and MTE to the outputs of all Attention and MLP blocks during the forward propagation using PyTorch’s Hook function. Then, during the backward propagation, we compute the gradient of the current adversarial example and use MIM \cite{mi} to update it.

\begin{table}[!t]
\centering
\setlength{\tabcolsep}{3mm}
\begin{tabular}{ccccc}
\toprule[1pt]
Model & $I$ & $d$ & $s$& $\eta$ \\ \midrule[1pt]
ViT-B  & 0,1,4,9,11  & 25 & 0.8 & 0.3 \\
CaiT-S & 2,14,25     & 30 & 0.6 & 0.2 \\
PiT-T  & 1,6,11      & 25 & 0.7 & 0.2 \\
DeiT-B & 0,1,5,10,11 & 30 & 0.7 & 0.2 \\ \bottomrule[1pt]
\end{tabular}
\caption{The optimal parameter settings of our FPR about its index set ($I$), diversity factor ($d$), scale factor ($s$), and attenuation factor ($\eta$). Detailed ablation studies in Section~\ref{ablsec} provide the sensitivity analysis of the attack performance towards the parameter selection under four cases.}
\label{parameter}
\end{table}

\begin{table*}[!htb]
\centering
\resizebox{\linewidth}{!}{
\begin{tabular}{cccccccccccccccc}
\toprule[1pt]
Model                   & Attack & ViT-B          & CaiT-S          & PiT-B         & Vis-S         & Swin-T        & DeiT-T        & CoaT-T        & RN-18         & VGG-16        & DN-121        & EN-b0         & MN-v3         & RNX-50        & Average       \\ \midrule[1pt]
\multirow{5}{*}{ViT-B}  & MIM \cite{mi}   & \textcolor{gray}{97.2}          & 61.8            & 40.4          & 42.3          & 54.9          & 65.8          & 44.4          & 51.7          & 57.2          & 51.3          & 50.8          & 51.7          & 38.2          & 50.9          \\
                        & PNAPO \cite{pna} & \textcolor{gray}{99.1}          & 83.2            & 62.1          & 65.8          & 74.7          & 83.0          & 64.0          & 67.4          & 70.0          & 67.6          & 68.5          & 63.0          & 56.3          & 68.8          \\
                        & TGR \cite{tgr}   & \textcolor{gray}{99.1}          & 86.2            & 63.8          & 69.9          & 81.1          & 94.4          & 70.4          & 79.1          & 79.8          & 75.3          & 83.1          & 79.9          & 60.0          & 76.9          \\
                        & GNS \cite{gns}   & \textcolor{gray}{99.9} & 89.5            & 68.1          & 74.0          & 82.4          & 93.6          & 72.2          & 77.8          & 76.7          & 76.0          & 79.1          & 76.0          & 62.0          & 77.3          \\
                        & FPR& \textcolor{gray}{99.2} & \textbf{92.5}   & \textbf{73.0} & \textbf{78.3} & \textbf{88.5} & \textbf{98.2} & \textbf{77.7} & \textbf{87.3} & \textbf{87.6} & \textbf{82.0} & \textbf{90.3} & \textbf{87.4} & \textbf{68.3} & \textbf{84.3} \\ \hline
\multirow{5}{*}{CaiT-S} & MIM \cite{mi}   & 68.5           & \textcolor{gray}{98.8}           & 50.0          & 54.9          & 68.2          & 75.3          & 54.8          & 61.0          & 67.8          & 59.8          & 61.0          & 55.7          & 47.1          & 60.3          \\
                        & PNAPO \cite{pna} & 69.8           & \textcolor{gray}{88.2}           & 58.1          & 61.0          & 70.3          & 73.8          & 60.1          & 64.8          & 69.7          & 64.3          & 65.5          & 61.7          & 53.2          & 64.4          \\
                        & TGR \cite{tgr}   & 90.1           & \textcolor{gray}{100.0} & 76.0          & 81.3          & 91.2          & 97.8          & 80.6          & 86.9          & 86.6          & 84.3          & 87.7          & 82.1          & 69.1          & 84.5          \\
                        & GNS \cite{gns}   & 91.2           & \textcolor{gray}{100.0} & 78.6          & 82.2          & 91.6          & 97.6          & 81.1          & 86.6          & 87.1          & 84.9          & 89.6          & 81.9          & 70.8          & 85.3          \\
                        & FPR& \textbf{93.8}  & \textcolor{gray}{100.0} & \textbf{81.6} & \textbf{85.6} & \textbf{95.0} & \textbf{98.8} & \textbf{86.8} & \textbf{87.4} & \textbf{88.7} & \textbf{88.6} & \textbf{90.6} & \textbf{82.9} & \textbf{76.8} & \textbf{88.1} \\ \hline
\multirow{5}{*}{PiT-T}  & MIM \cite{mi}   & 30.0           & 36.1            & 38.5          & 43.3          & 51.0          & 75.0          & 55.5          & 69.5          & 68.7          & 60.8          & 64.5          & 71.6          & 40.4          & 54.2          \\
                        & PNAPO \cite{pna} & 39.5           & 50.9            & 49.6          & 54.9          & 61.6          & 85.4          & 65.0          & 78.7          & 76.7          & 72.2          & 77.9          & 81.0          & 50.8          & 64.9          \\
                        & TGR  \cite{tgr}  & 37.9           & 47.7            & 46.9          & 54.1          & 62.5          & 87.4          & 65.1          & 82.8          & 82.0          & 75.5          & 81.8          & 88.8          & 53.7          & 66.6          \\
                        & GNS \cite{gns}   & 37.1           & 44.1            & 46.4          & 51.9          & 59.4          & 84.4          & 63.3          & 79.9          & 78.0          & 71.5          & 76.3          & 84.3          & 48.6          & 63.5          \\
                        & FPR& \textbf{44.3}  & \textbf{56.1}   & \textbf{57.1} & \textbf{62.4} & \textbf{69.5} & \textbf{91.3} & \textbf{72.2} & \textbf{88.7} & \textbf{86.4} & \textbf{80.4} & \textbf{87.6} & \textbf{91.1} & \textbf{60.3} & \textbf{72.9} \\ \hline
\multirow{5}{*}{DeiT-B} & MIM  \cite{mi}  & 86.0           & 82.4            & 67.0          & 69.5          & 79.7          & 87.7          & 65.3          & 69.0          & 70.5          & 70.1          & 71.4          & 61.4          & 58.2          & 72.2          \\
                        & PNAPO \cite{pna} & 83.4           & 87.7            & 72.4          & 74.1          & 82.2          & 86.0          & 71.9          & 74.9          & 74.7          & 74.9          & 76.1          & 68.0          & 66.3          & 76.4          \\
                        & TGR  \cite{tgr}  & 92.4           & 97.1            & 81.2          & 84.3          & 92.6          & 97.9          & 83.5          & 86.0          & 85.2          & 85.0          & 89.9          & 84.5          & 74.6          & 87.2          \\
                        & GNS \cite{gns}   & 92.1           & 98.0            & 80.5          & 84.2          & 91.0          & 98.0          & 79.8          & 83.3          & 80.7          & 83.5          & 86.7          & 81.3          & 70.8          & 85.4          \\
                        & FPR& \textbf{94.6}  & \textbf{98.4}   & \textbf{83.8} & \textbf{86.7} & \textbf{95.5} & \textbf{99.5} & \textbf{87.0} & \textbf{89.8} & \textbf{89.2} & \textbf{88.1} & \textbf{93.9} & \textbf{86.8} & \textbf{79.8} & \textbf{90.2} \\ \bottomrule[1pt]
\end{tabular}
}
\caption{Attack success rates ($\%$) of different surrogate refinements on various ViTs and CNNs. The “Average” column excludes the white-box results with grey font.}
\label{attack_normal}
\end{table*}

\section{Experiments}
\subsection{Experimental Settings}
\noindent\textbf{Datasets and baselines.}
Following the common practice \cite{vt,gra,tgr}, we use 1000 images randomly selected from the ILSVRC2012 \cite{image} validation set.
We compare our FPR with other surrogate refinements that focus on backward propagation: PNAPO \cite{pna}, TGR \cite{tgr}, and GNS \cite{gns}.
All these methods are built on MIM \cite{mi}.

\noindent\textbf{Models and defenses.}
For surrogate models, we consider four popular ViTs: ViT-Base (ViT-B) \cite{vit}, CaiT-Small (CaiT-S) \cite{cait}, PiT-Tiny (PiT-T) \cite{pit} and Deit-Base (Deit-B) \cite{deit}. For target models, we consider seven ViTs and six CNNs: ViT-B \cite{vit}, CaiT-S \cite{cait}, PiT-Base (PiT-B)~\cite{pit}, Visformer-Small (Vis-S) \cite{visf}, Swin-Tiny (Swin-T) \cite{swin}, DeiT-Tiny (Deit-T) \cite{deit}, and CoaT-Tiny (CoaT-T) \cite{coat}, ResNet-18 (RN-18) \cite{resnet}, VGG-16 \cite{vgg}, DenseNet-121 (DN-121) \cite{dense}, EfficientNet-b0 (EN-b0) \cite{eff}, MobileNet-v3 (MN-v3) \cite{mob}, and ResNeXt-50 (RNX-50) \cite{resnext}.
Five popular defenses are tested: adversarial training (AT) paired with RN-50 \cite{at},  high-level representation guided denoiser (HGD) \cite{hgd}, neural representation purifier (NRP) \cite{nrp}, Bit depth reduction (BDR) \cite{bit}, and JPEG compression \cite{jpeg}. 
Finally, two online models, i.e., Baidu Cloud API and Aliyun API, are also included.
					
\noindent\textbf{Hyperparameter settings.}
Commonly, we set the number of iterations as 10, the maximum perturbation bound $\epsilon$ as 16, the step size $\alpha$ as 1.6, and the decay factor of momentum gradient $\mu=1$. Parameters of all baselines are the same as their default values. 
For our FPR, the optimal hyperparameters are listed in Table~\ref{parameter}, with a detailed analysis in Section~\ref{ablsec} discussing their sensitivity to different parameter choices.
All experiments are conducted on a single RTX 4060 GPU with 8 GB of VRAM.

\subsection{Experimental Results}
\label{total_experiment}

\begin{table}[t]
\centering
\resizebox{\linewidth}{!}{
\begin{tabular}{cccccccc}
\toprule[1pt]
Model                   & Method & AT            & HGD           & NRP           & JPEG          & Bit           & Average       \\ \midrule[1pt]
\multirow{4}{*}{ViT-B}  & MIM \cite{mi}    & 43.9          & 29.8          & 42.1          & 57.5          & 59.1          & 46.5          \\
                        & PNAPO \cite{pna} & 45.6          & 47.5          & 53.5          & 70.9          & 74.8          & 58.4          \\
                        & TGR \cite{tgr}   & 49.7          & 50.8          & 61.6          & 77.7          & 82.0          & 64.4          \\
                        & GNS \cite{gns} & 49.2 	&54.8 	&61.3 	&77.9 	&82.3 	&65.1 \\
                        & FPR & \textbf{50.7} & \textbf{61.4} & \textbf{69.2} & \textbf{84.5} & \textbf{88.4} & \textbf{70.8} \\ \hline
\multirow{4}{*}{CaiT-S} & MIM \cite{mi}    & 44.3          & 37.4          & 46.2          & 64.1          & 66.9          & 51.8          \\
                        & PNAPO \cite{pna} & 44.8          & 43.7          & 48.2          & 65.2          & 67.3          & 53.8          \\
                        & TGR \cite{tgr}   & 49.8          & 60.7          & 66.4          & 81.2          & 86.8          & 69.0          \\
                        & GNS \cite{gns} &49.8 	&62.7 	&67.4 	&82.6 	&88.0 	&70.1 \\
                        & FPR  & \textbf{51.1} & \textbf{68.9} & \textbf{70.8} & \textbf{86.2} & \textbf{90.4} & \textbf{73.5} \\ \hline
\multirow{4}{*}{PiT-T}  & MIM \cite{mi}    & 46.1          & 32.0          & 42.0          & 57.5          & 62.0          & 47.9          \\
                        & PNAPO \cite{pna} & 47.5          & 42.9          & 48.5          & 65.5          & 69.8          & 54.8          \\
                        & TGR  \cite{tgr}  & 50.8          & 44.0          & 50.5          & 68.2          & 71.9          & 57.1          \\
                        &GNS \cite{tgr} &48.9 	&38.6 	&49.3 	&65.3 	&68.8 	&54.2 \\
                        & FPR & \textbf{51.4} & \textbf{49.9} & \textbf{56.1} & \textbf{73.2} & \textbf{77.4} & \textbf{61.6} \\ \hline
\multirow{4}{*}{DeiT-B} & MIM  \cite{mi}   & 44.7          & 48.1          & 52.2          & 70.3          & 74.9          & 58.1          \\
                        & PNAPO \cite{pna} & 46.9          & 57.2          & 57.0          & 73.6          & 78.7          & 62.7          \\
                        & TGR  \cite{tgr}  & \textbf{50.2} & 65.2          & 67.8          & 82.2          & 88.1          & 70.7          \\
                        &GNS \cite{gns} &48.3 	&61.6 	&65.7 	&80.6 	&86.3 	&68.5 
\\
                        & FPR & \textbf{50.2} & \textbf{70.7} & \textbf{71.3} & \textbf{87.0} & \textbf{91.5} & \textbf{74.1} \\ \bottomrule[1pt]
\end{tabular}
}
\caption{Attack success rates ($\%$) of different surrogate refinements against various defenses. AT is trained with RN-50; HGD is based on Inc-v3; NRP, JPEG and Bit are averaged on seven ViTs and six CNNs.}
\label{attack_defenses}
\end{table}
\begin{table}[t]
\centering
\setlength{\tabcolsep}{8.0mm}
\begin{tabular}{ccc}
\toprule[1pt]
Method    & ViTs          & CNNs          \\ \midrule[1pt]
SIA \cite{sia}      & 73.4          & 69.1          \\
SIA+FPR& \textbf{86.6} & \textbf{90.4} \\ \hline
BSR \cite{bsr}      & 69.4          & 66.9          \\
BSR+FPR& \textbf{84.8} & \textbf{89.9} \\ \hline
VTM \cite{vt}      & 63.1          & 57.3          \\
VTM+FPR& \textbf{84.0} & \textbf{81.7} \\ \hline
GRA  \cite{gra}     & 73.6          & 68.3          \\
GRA+FPR& \textbf{86.1} & \textbf{85.3} \\ \bottomrule[1pt]
\end{tabular}
\caption{Attack success rates ($\%$) of combined methods averaged on six CNNs and six ViTs (without surrogate model ViT-B).}
\label{attack_combination}
\end{table}

\noindent\textbf{Different target models.}
We first compare FPR with baselines on transferring adversarial examples from ViTs to different ViTs and CNNs. 
Table \ref{attack_normal} shows that FPR consistently outperforms baselines in all cases, suggesting the potential of refining the surrogate model from the perspective of forward propagation.
For example, when generating adversarial examples on ViT-B, FPR achieves an average attack success rate of 84.3\% on six CNNs, surpassing GNS, TGR, and PNAPO by 7.0\%, 7.4\%, and 15.5\%, respectively, and improves the baseline MIM by 33.4\%.

\noindent\textbf{Different defenses.}
We then evaluate different attack methods against five popular defenses.
As shown in Table \ref{attack_defenses}, FPR maintains its superiority over existing surrogate refinements in such challenging defense settings.
For instance, when generating adversarial examples on ViT-B, FPR achieves an average attack success rate of 70.8\% across five defenses, outperforming other surrogate refinements GNS, TGR, and PNAPO by 5.7\%, 6.4\%, and 12.4\%, respectively, while surpassing the vanilla baseline, MIM, by 24.3\%.

\noindent\textbf{Incorporating other transfer methods.}
A good surrogate refinement-based transfer method should also exhibit strong compatibility with other (types of) transfer-based attack methods. In this study, we integrate FPR into two advanced gradient correction-based transferable attacks, VTM \cite{vt} and GRA \cite{gra}, as well as two latest input augmentation-based transferable attacks, BSR \cite{bsr} and SIA \cite{sia}.
Specifically, we equip these four attacks with 5 augmented copies per iteration, while keeping other parameter settings as their default values. Results in Table \ref{attack_combination} indicate that FPR can effectively complement existing transfer-based attacks for stronger transferability. For instance, when combined with our FPR, the attack success rate of VTM increases from 57.3\% to 81.7\% on CNNs.

\noindent\textbf{Online models.} 
We also evaluate our FPR on two online models, i.e., Baidu Cloud API and Aliyun API. The results, presented in Appendix \ref{ap_online}, show that FPR still outperforms both TGR and GNS.

\begin{table}[t]
\centering
\setlength{\tabcolsep}{6mm}
\begin{tabular}{ccc}
\toprule[1pt]
Method  & ViTs          & CNNs          \\ \midrule[1pt]
MIM \cite{mi}     & 51.6          & 50.2          \\
Dropout & 55.8          & 58.9          \\
TGR \cite{tgr} & 77.6 & 76.2\\
GNS \cite{gns} & 80.0 & 74.6\\
\hline
AMD  & 82.1          & 78.4          \\
MTE     & 78.4          & 79.6          \\
FPR (AMD+MTE)   & \textbf{84.7} & \textbf{83.8} \\ \bottomrule[1pt]
\end{tabular}
\caption{Ablation studies of different FPR components with ViT-B as the surrogate model. All involved methods are built on MIM. Dropout is applied at the same index set as AMD.}
\label{ablations}
\end{table}

\begin{figure*}[!htb]
\centering
    \includegraphics[width=2.0\columnwidth]{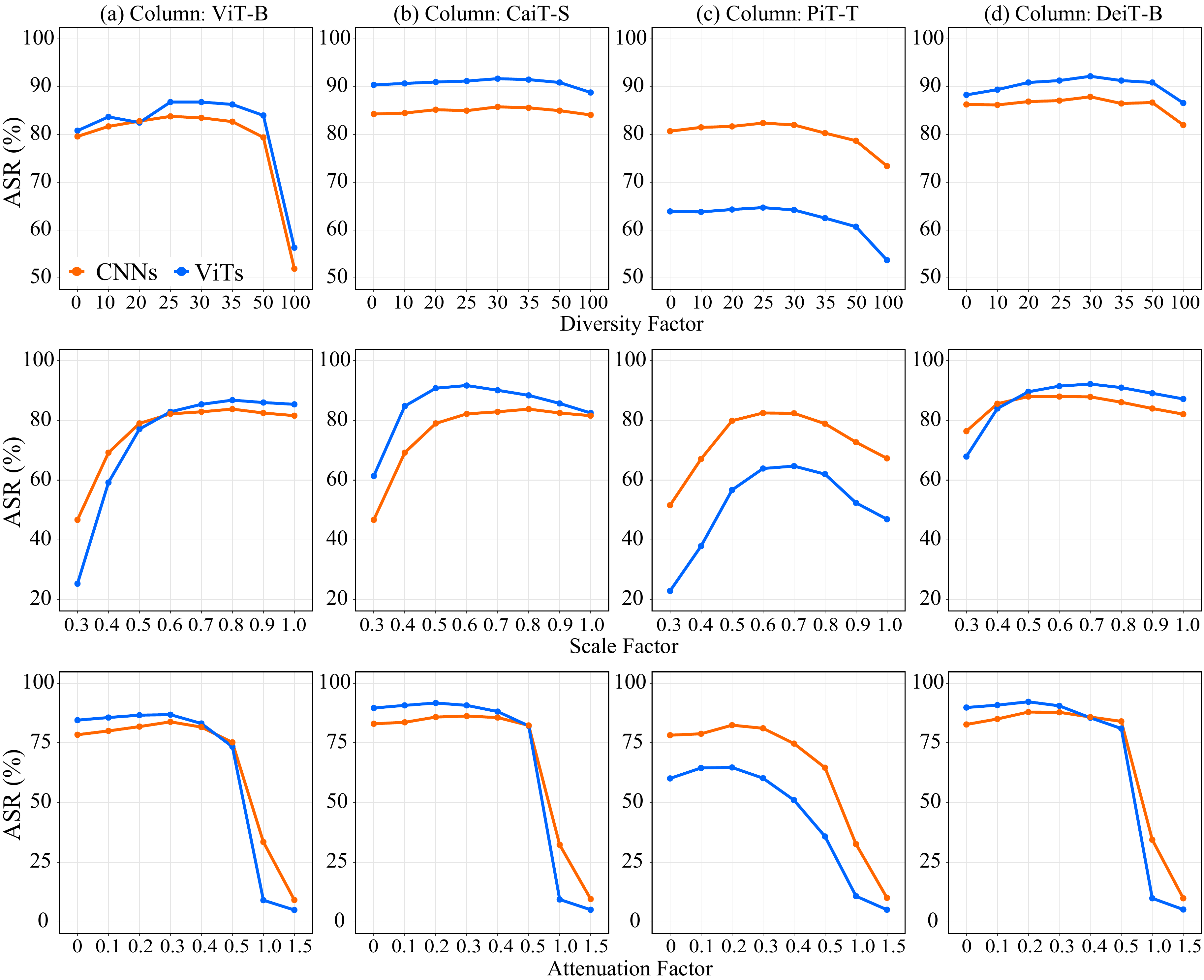}
\caption{Sensitivity analysis about diversity factor, scale factor, and attenuation factor, 
where ASR remains relatively stable within a certain range of values for these three factors.}
\label{sensitivity}
\end{figure*}

\subsection{Ablation Studies}
\label{ablsec}

\noindent\textbf{Individual components.} We compare each component of FPR with the plain baseline MIM, and the two best-performed surrogate refinements, TGR and GNS. We also consider Dropout as a comparison to our AMD.
From Table \ref{ablations}, we can see that combining our AMD and MTE yields the best transferability results. 
In addition, Dropout is less effective than our AMD, indicating that directly removing certain attention maps is rather aggressive and fails to retain useful information.

\noindent\textbf{Parameter sensitivity.} 
Figure \ref{sensitivity} shows the attack success rates when varying three main parameters: diversity factor $d$, scale factor $s$, and attenuation factor $\eta$. 
$d$ controls the diversity of the attention map. FPR achieves satisfactory performance when $d$ is between 25 and 30. 
$s$ represents the importance of current token embeddings. FPR achieves satisfactory performance when $s$ is between 0.6 and 0.8. 
$\eta$ signifies the importance of historical token embeddings. FPR achieves satisfactory performance when $\eta$ is between 0.2 and 0.3. 
Detailed results of applying the same index set ($I$) to other methods and the sensitivity of the $I$ used in AMD are reported in Appendix \ref{ap_index}, where we also provide insights for identifying the optimal $I$.

\noindent\textbf{Gradient vanishing comparison.}
We compare the extent of gradient vanishing in Table \ref{gv}. We can conclude that gradient vanishing can improve transferability, but it is not a simple linear correlation. This aligns with our claim that the key of AMD is to diversify attention maps in the forward propagation, while it implicitly and mildly induces some gradient vanishing in the backward propagation.

\begin{table}[!t]
\centering
\setlength{\tabcolsep}{0.06mm}
\small
\begin{tabular}{cccccc}
\toprule[1pt]
Attack  & MIM \cite{mi} & PNA \cite{pna} & TGR \cite{tgr} & GNS \cite{gns} & AMD \\ \toprule[1pt]
Mean & $2.9 \times 10^{-4}$ & 0.0 & $3.9 \times 10^{-10}$ & $2.4 \times 10^{-15}$ & $1.5 \times 10^{-5}$ \\
ASR & 50.9 & 60.6 & 63.2 & 62.7 & 80.3 \\ \toprule[1pt]
\end{tabular}
\vspace{-4mm} 
	\caption{
Mean of attention map's input gradient elements from 100 images over 10 iterations. All surrogate refinements adopt Index Set (0,1,4,9,11). ASR is averaged on 13 models and ViT-B is the surrogate model.}
	\label{gv}
\end{table}


\section{Conclusion}

In this work, we propose the Forward Propagation Refinement (FPR) to generate highly transferable adversarial examples on Vision Transformers (ViTs).
FPR is the first work to refine the surrogate ViTs during forward propagation entirely.
Specifically, FPR consists of two parts: Attention Map Diversification (AMD) and Momentum Token Embedding (MTE). AMD diversifies the attention maps and implicitly causes gradient vanishing to enhance transferability, while MTE accumulates historical token embeddings to stabilize current embeddings for more accurate updates. The superiority of FPR is supported by both empirical results and theoretical analyses.

\section*{Acknowledgments}
This research is supported by the National Key Research and Development Program of China (2023YFE0209800), the National Natural Science Foundation of China (62406240, 62376210, 62132011, 62206217, T2341003, U244120060, 62161160337, U2441240, U24B20185, U21B2018), the Shaanxi Province Key Industry Innovation Program (2023-ZDLGY-38). Thanks to the New Cornerstone Science Foundation and the Xplorer Prize.

{
    \small
    \bibliographystyle{ieeenat_fullname}
    \bibliography{main}
}


\appendix


\section{Attack on Online Models}
\label{ap_online}

Table \ref{online} presents the Attack Success Rates (ASRs) on two commercial APIs—Baidu Cloud API and Aliyun API—using MIM combined with three advanced surrogate refinement methods: TGR, GNS, and our FPR, evaluated on 50 originally correctly classified images. Among these, FPR attack demonstrates a clear advantage. Specifically, it achieves the highest ASR of 86\% on the Baidu Cloud API and 80\% on the Aliyun API, outperforming both TGR and GNS. This highlights the superior effectiveness of FPR in cheating commercial models.


\section{More Ablation Studies about Index Set}
\label{ap_index}

\subsection{Index Set among Different Methods}
For fairness, we conduct an ablation study on the index set $I$ and report the performance of various methods using the same index set. Table \ref{is_1} shows that both TGR and GNS experience a performance drop when using the index set from our method. Additionally, while the performance of AMD improves with the selected index set, it performs worse when applied to all blocks, as AMD makes more significant modifications to the  forward propagation. On the other hand, MTE is relatively mild in its modifications, which is why it performs better when applied to all blocks during forward propagation.

\subsection{Sensitivity of Index Set}
We analyze the sensitivity of index set 
used in AMD in Table \ref{is_2}, we can see that block selection is important since our optimal setting largely outperforms the random selection. 
In addition, the number of selected indexes also matters, since removing any block from our optimal setting slightly decreases the attack performance.

\subsection{How to Find the Optimal Index Set}

Based on our empirical findings, initially, we suggest starting from block 0 as the baseline. Then, AMD should be applied every 3-5 blocks, as this range tends to provide a balance between the overfitting (increrasing the diversity of attention map) and underfitting (introducing much unfamiliar information to the attention map). After this initial selection, we recommend fine-tuning the index set by replacing each current block with adjacent blocks, as this allows for more targeted adjustments while maintaining the diversity of the perturbations. This iterative refinement process helps in identifying the optimal index set for adversarial transferability.

\section{Visualization of Adversarial Examples}
\begin{figure}[t]
	\centering
	\includegraphics[width=1\columnwidth]{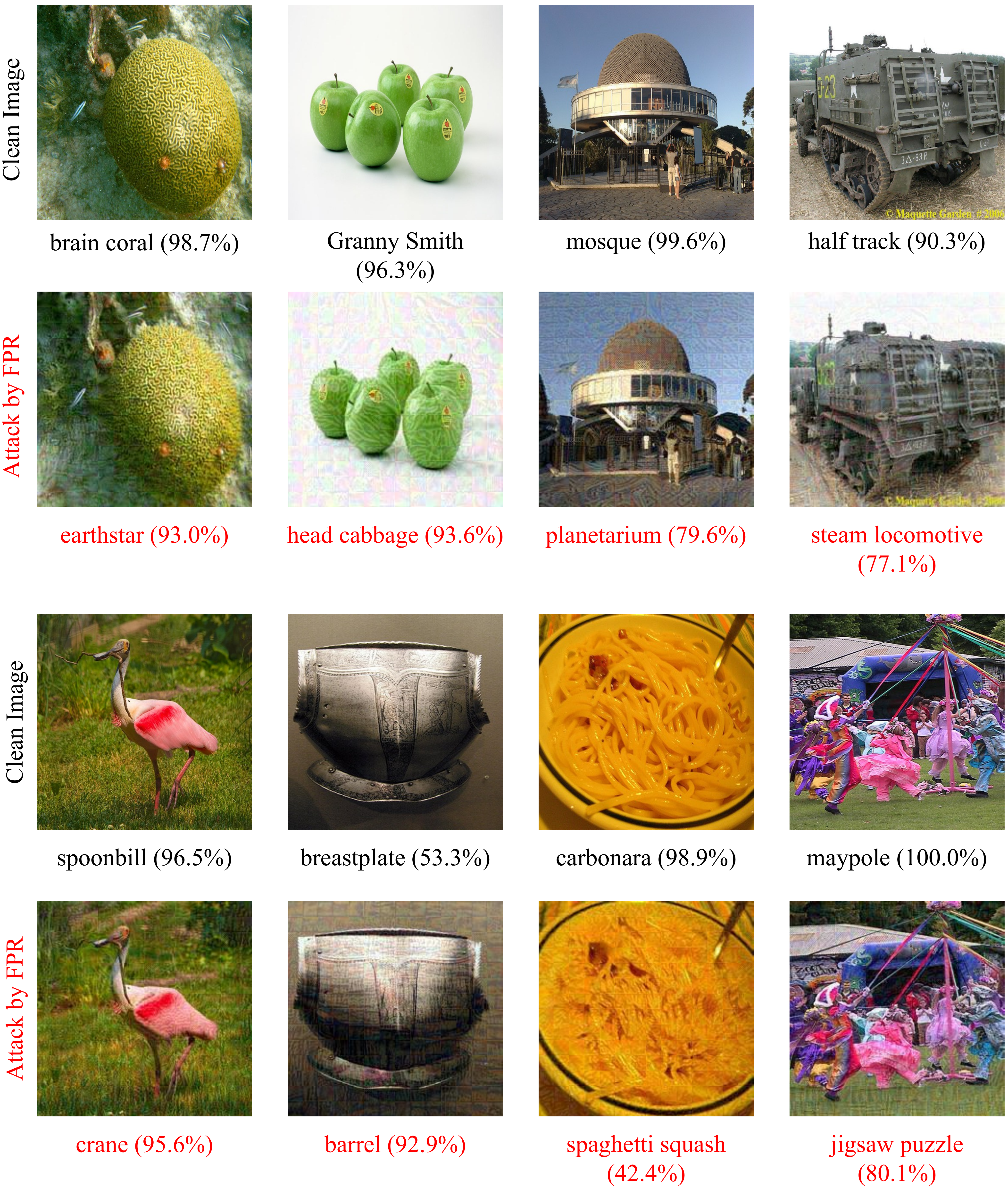}
	\caption{Visualizations for the crafted adversarial examples. Here the substitute model is ViT-B and the target model is RN-18.}
	\label{ap_vis}
\end{figure}

Figure \ref{ap_vis} provides the illustration of some crafted adversarial examples. It can be observed that, at first, the clean image can be classified with high confidence. However, their corresponding  adversarial examples can effectively fool the target models.

\begin{table}[!t]
\centering
\setlength{\tabcolsep}{3.5mm}
\begin{tabular}{ccc}
\toprule[1pt]
Attack (ViT-B) & Baidu Cloud API & Aliyun API \\ \toprule[1pt]
TGR & 0.78 & 0.78 \\
GNS & 0.76 & 0.74 \\
FPR & \textbf{0.86} & \textbf{0.80} \\ \toprule[1pt]
\end{tabular}
	\caption{
	ASRs on two commercial APIs are calculated on 50 originally correctly classified images.}
	\label{online}
\end{table}

\begin{table}[!t]
\centering
\setlength{\tabcolsep}{5.0mm}
\begin{tabular}{ccc}
\toprule[1pt]
Attack (ViT-B) & CNNs & ViTs \\ \toprule[1pt]
TGR (All, Default) & 76.2 & 77.6 \\
TGR (0,1,4,9,11) & 60.7 & 65.6 \\
GNS (All, Default) & 74.6 & 80.0 \\
GNS (0,1,4,9,11) & 60.5 & 64.9 \\ \hline
AMD (0,1,4,9,11, Default) & 78.4 & 82.1 \\
AMD (ALL) & 47.0 & 34.1 \\
MTE (All, Default) & 79.6 & 78.4 \\
MTE (0,1,4,9,11) & 59.8 & 63.6 \\
FPR  (AMD+MTE) & \textbf{83.8} & \textbf{84.7} \\ \toprule[1pt]
\end{tabular}
	\caption{
	ASRs on two commercial APIs are calculated on 50 originally correctly classified images.}
	\label{is_1}
\end{table}

\begin{table}[!htb]
\centering
\setlength{\tabcolsep}{3.2mm}
\begin{tabular}{cccc}
\toprule[1pt]
Model & Index Set & CNNs & ViTs \\ \toprule[1pt]
\multirow{7}{*}{ViT-B} & Random & 70.7 & 67.7 \\
 & 1,4,9,11 & 81.8 & 83.4 \\
 & 0,4,9,11 & 79.7 & 81.9 \\
 & 0,1,9,11 & 83.7 & 84.4 \\
 & 0,1,4,9 & 82.0 & 83.0 \\
 & 0,1,4,11 & 83.5 & 84.6 \\
 & 0,1,4,9,11 (Default) & \textbf{83.8} & \textbf{84.7} \\ \hline
\multirow{5}{*}{CaiT-S} & Random & 80.2 & 84.8 \\
 & 14,25 & 84.8 & 89.8 \\
 & 2,25 & 84.3 & 89.2 \\
 & 2,14 & 83.6 & 89.0 \\
 & 2,14,25 (Default) & \textbf{85.8} & \textbf{90.3} \\ \hline
\multirow{5}{*}{PiT-T} & Random & 76.0 & 55.0 \\
 & 6,11 & 81.1 & 63.1 \\
 & 1,11 & 81.3 & 64.6 \\
 & 1,6 & 81.1 & 64.4 \\
 & 1,6,11 (Default) & \textbf{82.4} & \textbf{64.7} \\ \hline
\multirow{7}{*}{DeiT-B} & Random & 85.9 & 89.3 \\
 & 1,5,10,11 & 85.9 & 91.0 \\
 & 0,5,10,11 & 87.0 & 91.0 \\
 & 0,1,10,11 & 87.0 & 91.4 \\
 & 0,1,5,11 & 87.7 & 92.2 \\
 & 0,1,5,10 & 87.5 & 92.0 \\
 & 0,1,5,10,11 (Default) & \textbf{87.9} & \textbf{92.2} \\ \toprule[1pt]
\end{tabular}
\caption{
Sensitivity analysis about the index set used in AMD.
``Random'' refers to randomly selecting the same number of blocks as in AMD.}
\label{is_2}
\end{table}

\end{document}